\documentclass[final]{elsarticle}
\usepackage{lineno,hyperref}
\usepackage{graphicx}        
\usepackage{multicol}        
\usepackage[bottom]{footmisc}
\usepackage{graphicx}      

\usepackage{pstricks}
\usepackage{graphics} 
\usepackage{graphicx}
\usepackage{caption}
\usepackage{nicefrac}
\usepackage{algpseudocode}
\usepackage{algorithm}
\usepackage[]{units}
\usepackage{paralist}
\usepackage{mathtools}

\usepackage{xcolor}
\usepackage{siunitx}
\usepackage{multirow}
\usepackage{pbox}

\usepackage{url}
\usepackage{epstopdf}
\graphicspath{ {Pictures/} }

\usepackage{epsfig}
\usepackage{float}
\usepackage{caption}
\usepackage{amsmath} 
\usepackage{amssymb}  
\usepackage{mathtools}
\graphicspath{ {Pictures/} }
\DeclareGraphicsExtensions{.pdf,.eps}
\usepackage{lipsum}

\newcommand*{\smallmat}[1]
  {\left[\begin{smallmatrix}#1\end{smallmatrix}\right]}
  
\begin{document}


\begin{frontmatter}
\title{Visual Area Coverage with Attitude-Dependent Camera Footprints by Particle Harvesting \tnoteref{mytitlenote}}
\tnotetext[mytitlenote]{This work has been partially funded by the European Union Horizon 2020 Research and Innovation Programme under the Grant Agreement No.644128 - AEROWORKS and the Grant Agreement No. 730302  (SIMS).}
\author[First]{Sina Sharif Mansouri} 
\author[Second]{Pantelis Sopasakis} 
\author[Third]{George Georgoulas} 
\author[First]{Thomas Gustafsson}
\author[First]{George Nikolakopoulos}
\address[First]{Robotics Group, Control Engineering Division, Department of Computer, Electrical and Space Engineering, Lule\r{a} University of Technology, Lule\r{a} SE-97187, Sweden, Lule\r{a} SE-97187, Sweden, Emails: \texttt{\{sinsha,tgu,geonik\}@ltu.se}}
\address[Second]{Queen's University Belfast, School of Electronics Electrical Engineering and Computer Science (EEECS) and Centre for Intelligent and Autonomous Manufacturing (i-AMS), Belfast, Northern Ireland, UK,  Email: \texttt{p.sopasakis@qub.ac.uk}}
\address[Third]{Senior Engineer, DataWise Data Engineering LLC,  Email: \texttt{gerorge.georgoulas@datawise.ai}}
\begin{abstract}  
In aerial visual area coverage missions, the camera footprint changes over time based on the camera position and orientation --- a fact that complicates the whole process of coverage and path planning. This article proposes a solution to the problem of visual coverage by filling the target area with a set of randomly distributed particles and harvesting them by camera footprints. This way, high coverage is obtained at a low computational cost. In this approach, the path planner considers six degrees of freedom (DoF) for the camera movement and commands thrust and attitude references to a lower-layer controller, while maximizing the covered area and coverage quality. The proposed method requires \textit{a priori} information of the boundaries of the target area and can handle areas of very complex and highly non-convex geometry. The effectiveness of the approach is demonstrated in multiple simulations in terms of computational efficiency and coverage.
\end{abstract}
\end{frontmatter}

\section{Introduction}
\subsection{Problem Statement}
Unmanned aerial vehicles (UAVs) equipped with visual sensors are rapidly emerging as the solution of choice for gathering information in many applications, especially in hostile or challenging environments, such as sensing of landslides~\cite{niethammer2010uav}, search and rescue missions~\cite{doherty2007uav} and forest fire inspection~\cite{alexis_coordination_2009}. Furthermore, UAVs are capable of providing high-resolution data/images that can be analyzed and used to investigate area characteristics, produce sparse and dense surface reconstructions~\cite{mansouri2018cooperative}, as well as hazard maps.

However, UAVs are subject to limitations related to payload and flight time~\cite{Sinabattry2017}. Payload limitations can prevent the usage of certain visual sensors and have a direct impact on flight time, thus lightweight cameras are preferred. Moreover, time limitations need to be compensated by optimization-based path planners, which aim at minimizing total flight length and duration.



\begin{figure}[htbp]
\centering
\includegraphics[width=0.65\linewidth]{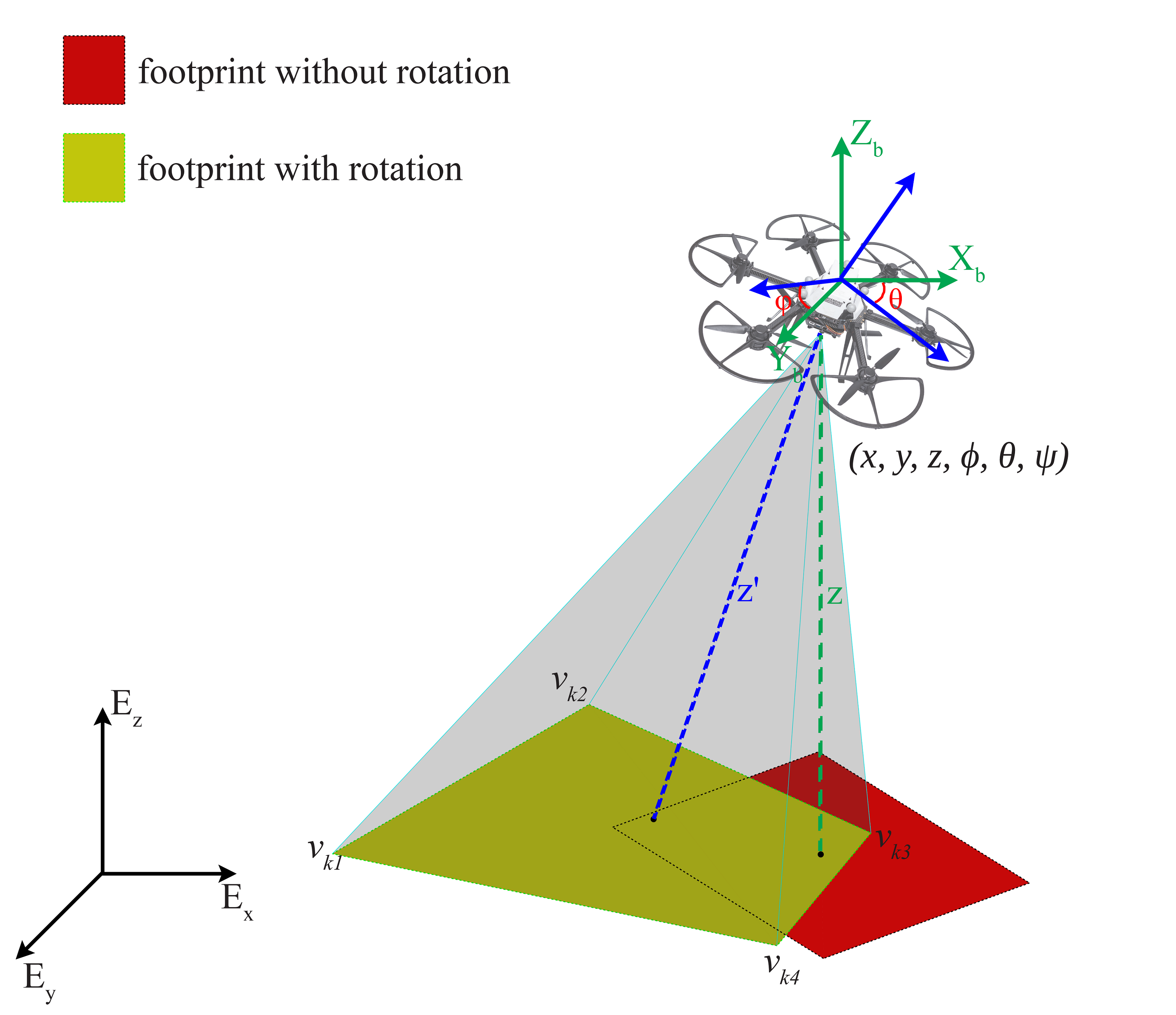}
\caption{Schematic of the field of view in the case of UAV with body and fixed frames. The red rectangle is the camera footprint when the UAV is at an upright position. The yellow rectangle is the camera footprint considering the tilt of the UAV. The two footprints can be significantly different.}
\label{fig:field}
\end{figure}

In the presented approach, the UAV can maneuver in six degrees of freedom (DoF) ($x,y,z, \phi, \theta, \psi$), while having a downward-looking camera fixed to the UAV frame. In this case, the camera footprint can be modeled by a polygon with four vertices $(v_{k1},v_{k2},v_{k3},v_{k4})$ called a \textit{cell} $c_k$ at time instant $k\in\mathbb{N}$. The size and shape of a cell varies with the position and orientation of the UAV. The general concept is illustrated in Fig.~\ref{fig:field} where the red and yellow cells are the camera footprints of the UAV at an upright and a tilted attitude. 

As one may observe, the covered area is different when the UAV deviates from the upright forward-facing orientation, while the description of the overall problem statement will be presented in Section~\ref{ProblemStatement}. Disregarding the fact that the camera footprint depends on the attitude of the UAV may lead to insufficient coverage and suboptimal behavior. Thus, the main objective of this article is to propose a path planner while maximizing coverage. The proposed method takes into account the target area and generates attitude and thrust commands, while considering all DoF of the UAV/camera, resulting in a time-varying and attitude dependent camera footprint.

\subsection{Background \& Motivation}
The 2D coverage path planning (CPP) problem, has been extensively investigated~\cite{galceran2013survey,raffo2011path}. 
Most CPP methods rely on the decomposition of the target area in sub-regions, which the agent subsequently sweeps~\cite{torres2016coverage,perez2016selecting,conesa2016mix}. A common taxonomy of CPP algorithms is based on the type of decomposition~\cite{galceran2013survey}: a) exact cellular decomposition methods~\cite{choset1998coverage} where the area is broken down to simple non-overlapping regions which the UAV traverses, b) Morse-based cellular decomposition methods~\cite{acar2002morse} where the area is decomposed based on critical points of Morse functions~\cite{milnor2016morse} and a motion planner algorithm guarantees that the agent passes through the critical points in the target area, c) landmark-based topological coverage methods where natural landmarks guide the decomposition process~\cite{wong2006qualitative}, d) contact sensor-based coverage of rectilinear environments~\cite{butler1999contact}, where the vehicle follows a cyclic path while building up a cellular decomposition of the area, e) grid-based methods~\cite{shivashankar2011real} where the target area is decomposed into a collection of uniform grid cells, and f) graph-based methods~\cite{xu2011graph} where the environment is represented by a graph, which is updated using available UAV sensors, while performing a coverage task. 

However, none of the aforementioned methods considers the sensors' placement and alignment on the UAV and the changing shape and size of the camera footprint with the attitude of the UAV. Not taking into consideration these factors can lead to low visual feedback quality or result in an unsatisfactory coverage outcome. Moreover, in most of the related approaches, the agent is considered to maintain a constant altitude and attitude of flight, while the segmentation is performed by a top-level procedure~\cite{valente2011multi,avellar2015multi}. Few works consider a three DoF movement and a camera footprint in path planning, e.g., in~\cite{popovic2017online} the informative path planning algorithm is proposed. In that approach, the global viewpoint selection and evolutionary optimization are combined to refine the UAVs trajectory. However, that approach is minimizing the path without taking into account of the camera orientation. Similarly, in~\cite{MANSOURI20181}, the target area is decomposed in relation to the camera footprint and the shortest path, which pass through all the camera footprints, is generated. However, the UAV altitude and attitude (roll and pitch) are assumed to be constant and only three DoF ($x,y,\psi$) are considered, while that is an off-line method that requires high computation power.

To the best of our knowledge no work so far has considered path planning taking into account the motion and orientation of the camera with six DoF, which results in non-constant attitude-dependent camera footprints. 

\subsection{Contributions}
Based on the aforementioned state of the art, the main contribution of this article is two-fold. The first major contribution, we address the problem of coverage of a path planner coupled with a time-varying attitude-dependent camera footprint for the first time, to the best of our knowledge. In this approach, the downward camera on the UAV possesses six DoF. The optimization procedure, aims at maximizing the covered area, and the coverage quality and provides input to the low level attitude controller. To provide close to real time computation for this highly complex problem of path planning instead of calculating the union of camera footprint cells and intersection of them to the target area, the target area is filled by a set of randomly generated particles and the path planner tries to harvest all particles by camera footprint cells. At each time instant the area is updated based on the previous covered area, points from visited areas are removed and the cells are calculated based on the position and orientation of the camera. Moreover, the UAV can maneuver in all directions and changes its altitude, this can effect the quality of visual feedback. As the UAV increases the altitude, the larger area is covered which results to shorter path however, the quality of the coverage is reduced. Thus, coverage quality function is proposed and added to the optimization problem; the term calculates the quality of the coverage based on the altitude and attitude of the UAV. 

The second major contribution stems from the fact that the segmentation of the area and the path planner are integrated, thus establishing an overall framework for the path planning which considers changes in the shape and size of the camera footprint of the UAV with a downward camera, which results to reasonably smooth and shorter paths as there is no need to sweep the area for full coverage.

The efficiency of the proposed method is evaluated on multiple case studies. In the proposed approach the camera movement and orientation are coupled with the UAV, while for cameras with gimbals, the camera remains horizontal regardless of the motion around them, a fact that limits the DoF of the camera motion in relation to the UAV. 

\subsection{Outline}
The rest of the article is structured as follows. Initially some mathematical preliminaries of the proposed problem are presented in Sections~\ref{UAV Kinematics} and~\ref{Coverage with Camera Cells} followed by the presentation of the coverage quality function in Section~\ref{Coverage quality function}. The proposed path planner is presented and discussed in Section~\ref{Path Planner}, while in Section~\ref{results} several simulation results are presented with a corresponding comparison and discussion. Finally Section~\ref{conclusion} concludes the article by summarizing the findings and offering some directions for future research.

\section{Problem Statement} \label{ProblemStatement}
\subsection{UAV Kinematics}\label{UAV Kinematics}
In order to describe the UAV kinematics, coordinate frames need to be defined. One coordinate system is fixed to the aerial vehicle and it is called the body frame $(X_b, Y_b, Z_b)$, while the other one is called the fixed (or inertial) frame and it is fixed to the earth $(E_x, E_y, E_z)$. The schematic structure of the UAV with different coordinate systems is illustrated in Fig.~\ref{fig:field}, where $z$ is the altitude of the UAV and $z'$ is the distance from the center of the UAV to the target area.
The state of the system is $X=[x,\dot{x}, y, \dot{y}, z, \dot{z}]^\top$, where $x$, $y$ and $z$ define the position of the UAV's center of mass and follow the following dynamics.
\begin{subequations}\label{eq:uavmodel}
\begin{align}
\ddot{x}&=\frac{T}{m}(\cos\psi \sin\theta \cos \phi+\sin \psi \sin \phi)\\
\ddot{y}&=\frac{T}{m}(\sin\psi \sin\theta \cos\phi-\cos\psi \sin\phi)\\
\ddot{z}&=-g+\frac{T}{m}(\cos\theta \cos\phi)
\end{align}
\end{subequations}
where $T$ is the total thrust which is exercised by propellers, $m$ is the mass of the system, $g$ is the gravity acceleration and $\phi, \theta$ and $\psi$ are the pitch, roll, and yaw of the UAV's attitude. More details about the modeling part can be found in~\cite{alexis2011switching}.
\subsection{Coverage with Camera Cells}\label{Coverage with Camera Cells}

Let us assume $\Omega \subset \mathbb{R}^2$ is a given region to be visually covered and the cells $c_k$ represent the camera's footprint at the $k^{th}$ $(k\in \mathbb{N})$ time instant. The goal is to generate attitude and thrust commands in order to maximize the coverage of the cells $c_k$ on the target area $\Omega$. In other words, the union of cells over time should cover the whole target area:
\begin{equation}
\Omega \subseteq  \bigcup\limits_{k \in \mathbb{N}} c_k (x_k,y_k,z_k,\theta_k,\phi_k,\psi_k)
\end{equation}
In this approach, the attitude and total thrust of the UAV is calculated by the path planner to maximize the covered area and the quality of the coverage. 
To be more specific, each cell $c_k$ can be identified by its vertices' positions $v_{kj}=(x_{kj},y_{kj},0),~j=\{1,2,3,4\}$. Naturally, the $z$-coordinate of $v_{kj}$ is zero because we need to cover an area at zero altitude but the proposed framework allows to accommodate uneven and non-horizontal surfaces. The vertices' positions obtained from the camera position $x_k,y_k,z_k$, camera orientation $\phi_k, \theta_k, \psi_k$, horizontal field of view (HFOV) $\alpha \in (0,180^\circ)$ and vertical field of view (VFOV) $\beta \in (0,180^\circ)$ of the camera~\cite{MANSOURI20181}. The $\alpha$ and $\beta$ are constants and can be obtained from camera specifications.

Furthermore, the orientation of the vehicle is represented by a rotation matrix. The three elementary rotation matrices $R_x$, $R_y$ and $R_z$, which rotate vectors by angles $\theta_k$, $\phi_k$ and $\psi_k$ about the $x$, $y$ and $z$ axes respectively, are:
\begin{subequations}
\label{eq:rotationmatrices}
\begin{align}
R_x=&\smallmat{
\phantom{\sin}1 & \phantom{\sin}0 &0 \\
\phantom{\sin}0 & \phantom{--}\cos\phi_k& -\sin\phi_k \\
\phantom{\sin}0&  \phantom{--}\sin\phi_k& \cos\phi_k
} \\
R_y=&\smallmat{
\cos\theta_k & \phantom{\sin}0 &\phantom{\sin}\sin\theta_k\\
    0 & \phantom{\sin}1&  \phantom{\sin}0\\
    -\sin\theta_k & \phantom{\sin}0 &\phantom{\sin}\cos\theta_k
} \\
R_z=&  \smallmat{
\cos\psi_k &-\sin\psi_k & \phantom{\sin}0\\
    \sin\psi_k &\cos\psi_k & \phantom{\sin}0\\
\phantom{-}     0 &  \phantom{-}0& \phantom{\sin}1
}
\end{align}
\end{subequations}

As it is shown in Fig.~\ref{fig:field}, the red rectangle is the camera footprint without consideration of the UAV's orientation. The vertices of the rectangle centered at $(0,0,0)$ can be calculated from the position of the UAV, and camera's VFOV and HFOV. In order to obtain the camera's footprint resulting from the orientation of the UAV, the vertices of the rectangle should be rotated by the rotation matrices and translated to the UAV's position. Therefore, the cells can be calculated by:
\begin{subequations}
\label{eq:verticesofcell}
\begin{align}
v_{k1} &= 
  \smallmat{
    x_k+R_{xyz}z_k\cot\alpha/2  \\
    y_k+R_{xyz}z_k\cot\beta/2  \\
0
} \\
v_{k2} &= \smallmat{
x_k+R_{xyz}z_k \cot\alpha/2 \\
y_k-R_{xyz}z_k \cot\beta/2 \\
0
}
\\
v_{k3} &=\smallmat{
x_k-R_{xyz}z_k \cot\alpha/2 \\
y_k-R_{xyz}z_k \cot\beta/2 \\
0
}
\\
v_{k4} &= \smallmat{
x_k-R_{xyz}z_k \cot\alpha/2 \\
y_k+R_{xyz}z_k \cot\beta/2 \\
0
}
\end{align}
\end{subequations}
where $(v_{k1},v_{k2},v_{k3},v_{k4})$ are the vertices of the cell at time $k$, $R_{xyz}=R_xR_yR_z$ and $(x_k,y_k,z_k)$ is the camera position at time $k$. As a remark, the area to be covered may not be flat (a subset of $\mathbb{R}^2\times\{0\}$). In such cases, the proposed approach is still applicable provided that the angle of visual perception is not of importance. This is the case when there are not any too high objects such as buildings, trees or hills that may obstruct the coverage.


\subsection{Image Quality} \label{Coverage quality function}
During the inspection, the UAV can maneuver in six DoF and fly at different altitudes. One solution for decreasing the length of the path is to increase the altitude of the UAV and allow higher tilts, which results to larger covered area. However, this solution results to images of lower resolution and reduces the quality of the coverage. Therefore, to take in account the quality of images in relation to the altitude of the UAV for the path planner, the coverage quality function $[z_{\min}, z_{\max}] \rightarrow [0, 1]$ is introduced as in~\cite{papatheodorou2017collaborative} (See Eq. \eqref{eq:qualityfunction} below). Without loss of generality, it is assumed that the coordinates of the camera $(X_c,Y_c,Z_c)$ are the same as those of the UAV and the distance from the camera to the center of mass of the UAV is negligible. The coverage quality is given by:
\begin{equation} \label{eq:qualityfunction}
q(z_k)= 
\begin{cases}
\frac{((-z_k-z_{\min})^2-(z_{\min}-z_{\max})^2)^2}{(z_{\min}-z_{\max})^4}      & \, \text{if } z_k \in [z_{\min}, z_{\max}]\\
0  &  \,   \text{otherwise}
\end{cases}
\end{equation}
where $q(z_k)=1$ and $q(z_k)=0$ correspond to the  minimum and maximum altitudes respectively. 


Additionally, in order to consider the tilt of the camera, the image quality is evaluated at the actual distance $z'$ between the camera and the area along the shooting direction. In case $\phi_k=0$ and $\theta_k=0$, $z_k$ and $z'_k$ will be equal, yet, nonzero values for $\phi$ and $\theta$ are necessary for the movement of the UAV, as it can be seen from ~\eqref{eq:uavmodel}.
 
The $z_k'$ distance increases with an increase in $\phi_k$ and $\theta_k$ at constant $z_k$. This means that the distance between the camera and the area can vary based on the orientation of the UAV. The corrected distance is given by:
\begin{equation} \label{eq:z_new}
z_k'=\frac{z_k}{\cos\phi_k \cos \theta_k}
\end{equation}    
By virtue of~\eqref{eq:z_new}, the coverage quality is evaluated as $q(z')$. 

\section{Path Planning by Particle Harvesting} \label{Path Planner}
\begin{figure} \centering
\includegraphics[width=\linewidth]{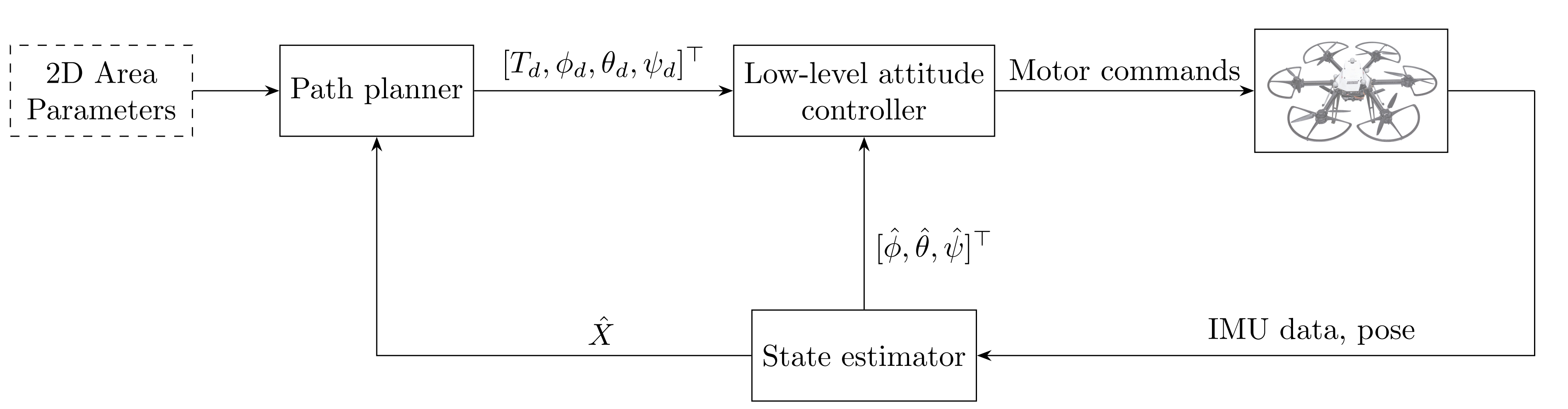}
\caption{Controller structure of the path planner. The path planner receives an estimate of the UAV's position from a state estimator and commands references of thrust and Euler angles to a lower-layer controller (typically an LQR). This separates the attitude control loops from the navigation control loop.}
\label{fig:ControlStructure}
\end{figure}

The path planner's objective is to generate attitude and thrust commands taking into account the camera footprint while trying to capture high quality images by avoiding high altitudes and high tilts. A block diagram representation of the proposed path planner and the corresponding low-level controller is shown in Fig.~\ref{fig:ControlStructure}. The path planner will generate a desired thrust $T_d$ and a desired orientation $\phi_d, \theta_d, \psi_d$ for the inner loop controller. The low-level attitude controller~\cite{XIONG2014725} is able to track the desired roll, pitch and yaw and to calculate the corresponding rotor speeds for the UAV. A multi-sensor-fusion extended Kalman filter (MSF-EKF)~\cite{msf} fuses the obtained pose information from the localization systems and the inertial measurement unit (IMU) measurements to provide orientation and position estimates of the vehicle.
The estimated position and velocity are provided to an outer control loop (the navigation controller) which controls the position and velocity of the UAV by providing references to an inner loop. The inner loop controller (the attitude controller), uses the attitude and angular velocity estimates to control the UAV's attitude.


Given the highly complex nature of the path planning problem, a formulation is proposed, which is based on filling the target domain $\Omega$ with a set of particles (points in \(\Omega\)) $\mathcal{I}_0$, generated by randomly sampling from a uniform distribution over $\Omega$. Then, the path planner harvests all particles in $\mathcal{I}_0$ by commanding thrust and tilt signals $u=[T, \phi, \theta, \psi]^\top$ to the inner loop controller (low-level controller), while taking into account the system dynamics and constraints. 

The state of the system is $X=[x,\dot{x},y,\dot{y},z,\dot{z}]^\top$, $\hat{X}$ is the estimated state, and $f$ is a discretization of~\eqref{eq:uavmodel}, which may be obtained by the Euler method, a Runge-Kutta method, or any other discretization method~\cite{stetter1973analysis}. Moreover, $X_{k+j|k}$ and $u_{k+j\mid k}$ are the state and control action ahead of $k+j$ steps form the current time $k$ and $\Delta u_{k+j|k}$ is difference of control action between $k+j-1$ and $k+j$ steps ahead of current time $k$.

For the path planner presented in Section~\ref{ProblemStatement} the following finite horizon cost function is introduced.
\begin{align}\label{eq:cost-function}
J &= \sum_{j=0}^{N-1} \underbrace{ w_X \|\Delta X_{k+j|k}\|^2}_\text{minimizing the movement} + \underbrace{w_I |\mathcal{I}_{k+j|k} \setminus c_{k+j|k}|} _\text{minimizing the remaining area}
  +\underbrace{(-w_q q(z'_{k+j|k})) }_\text{maximizing the quality}\notag\\  + &\underbrace{w_u \|\Delta u_{k+j|k}\|^2}_{\text{smoothness of control actions}} + \underbrace{w_z [z_{\min}-z_{k+j|k}]_+}_{\text{soft constraint on minimum altitude}}
\end{align}

The cost function of the optimization problem involves five terms. The first term minimizes the movement of the UAV which means the next position and orientation of the UAV should be close to the current one. This way the UAV avoids jerks. The second term minimizes the remaining area by removing the observed points from target area. The operation $|{I}_k \setminus c_k|$ denotes the cardinality of the set ${I}_k \setminus c_k$ and the operator $\setminus$ denotes the relative complement and is defined as follows:
\begin{equation}
\mathcal{I}_k \setminus c_k {}\mathrel{:}\!\mathop{=}{} \left\{x\in \mathcal{I}_k {}\mid{} x \not\in c_k\right\}
\end{equation}
In the third term of \eqref{eq:cost-function} the quality of the visual feedback is considered. Large values of $q(z'_{k+j|k})$ lead to higher coverage quality. The fourth term, $\|\Delta u_{k+j|k}\|^2$ penalizes the aggressiveness of the obtained control actions. The fifth term is the soft constraint on minimum hovering altitude of the UAV to reduce the ground effect disturbance.  What is more, altitude constraints guarantee that the UAV will not fly at too high an altitude, which is a safety as well as legal requirement\footnote{The EU parliament are discussing laws that will regulate the maximum altitude for civilian drones. Already all EU countries have in place regulations for civilian drones prohibiting flights above a certain altitude (typically 100m)~\cite{EUregulation}.}  The ground effect is the change in the thrust generated by the rotors when UAV is close to the ground due to the interaction of the rotor airflow with the ground surface~\cite{sanchez2017characterization}. The operation $[{}\cdot{}]_+$ is the plus operator which is defined as:
\begin{equation}
[{}z{}]_+=\max\{z,0\}
\end{equation}
Additionally, $w_X$, $w_I$, $w_q$, $w_u$ and $w_z$ are the weights for each term of the objective function which reflect the relative importance of each term; the highest importance is on $w_I$ to enforce high coverage. 

The following optimization problem is defined:
\begin{subequations}\label{eq:optimizationcoverage}
\begin{align}
&\underset{
  \{  u_{k+j\mid k} \}_{j=0}^{N-1}} {\mathbf{minimize}}\  J\\
&\mathbf{s.t.}\  
X_{k+j+1\mid k}=f(X_{k+j\mid k},u_{k+j\mid k}),\ \text{for}\, j\in \mathbb{N}_{[0,N-1]} 
\label{eq:optimizationcoverage:sysDyn}
\\ 
&z'_{k+j\mid k}=\frac{z_{k+j\mid k}}{\cos\phi_{k+j\mid k} \cos \theta_{k+j\mid k}},\ \text{for}\, j\in \mathbb{N}_{[0,N-1]} 
\\ 
&u_{k+j\mid k} \in [u_{\min},u_{\max}],\ \text{for}\, j\in \mathbb{N}_{[0,N-1]} 
\label{eq:optimizationcoverage:actuationConstraints}
\\ 
&z'_{k+j\mid k} \in [0,z'_{\max}],\ \text{for}\, j\in \mathbb{N}_{[1,N-1]} 
\label{eq:optimizationcoverage:altConstraints}
\\ 
&X_{k\mid k}=\hat{X}_k\\
&\mathcal{I}_{k\mid k} = \mathcal{I}_k
\end{align}
\end{subequations}
At every time instant $k$, a finite-horizon optimal path with prediction horizon $N$ is solved and a corresponding optimal sequence of control actions $u_{k|k}^{\star},\dots$ $u_{k+N-1|k}^{\star}$ are generated. The first control action $u_{k|k}^\star$ in that sequence is applied to the system in a receding horizon fashion as shown in Alg.~\ref{alg:pathplanner}, and the covered points are removed from $\mathcal{I}_k$. This way, the remaining area which can be covered along the prediction horizon is minimized and the path planner aims at maximizing coverage. The optimal control problem is solved in a receding horizon fashion~\cite{rawlings2009model}. Eventually, the target area is approximately covered and the mission comes to an end.
Note that the problem is solved subject to the system dynamics \eqref{eq:optimizationcoverage:sysDyn}, actuation constraints \eqref{eq:optimizationcoverage:actuationConstraints} and the altitude constraints \eqref{eq:optimizationcoverage:altConstraints}.

Furthermore, to calculate if the particles located inside of the cell, the ray casting algorithm~\cite{514556} is used. The ray casting method is a concept from computational geometry that determines if a point lies inside a bounded region (cell). The method counts the number of times that the ray casting from each point to a point outside of the cell intersects the cells boundaries. If the number of intersections is 0 or an even number, the point lies outside the cell otherwise lies inside. Moreover, the sequential quadratic programming method~\cite[Chap.~18]{nocedal2006sequential} of MATLAB's optimization toolbox (\texttt{fmincon}) running on a single core was used to solve the optimization problem. Overall, Alg.~\ref{alg:pathplanner} summarizes the steps of the proposed path planner.
\begin{algorithm}
\caption{Navigation algorithm for visual coverage with the proposed particle harvesting method.}
\label{alg:pathplanner}
\begin{algorithmic}[1]
\Require Target area $\Omega$, initial position $X_0$
\State Select random points $\{p_i\}_{i=1,\dots,N}$ from $\Omega$
\State $\mathcal{I}_0\leftarrow\{p_1,\dots,p_N\}$, $k\leftarrow 0$
\For{$k=0,\ldots$ and while $\mathcal{I}_k \neq \varnothing$}{ }
\State Obtain estimate $\hat{X}_k$ and compute $c_k$ using~\eqref{eq:verticesofcell}
\State  $\mathcal{I}_k \leftarrow \mathcal{I}_k \setminus c_k$
\State Solve~\eqref{eq:optimizationcoverage}, compute $u_k|k^\star$
       and apply it to the system       
\EndFor
\end{algorithmic}
\end{algorithm}


\section{Simulation Results}\label{results}
The proposed method has been evaluated in a simulation environment. A validated model of the Ascending Technologies NEO hexacopter has been selected for the simulations. 
The overall mass of NEO considered in this work is 3.3~$\unit{kg}$ and the total thrust is constrained between 0 and 50~$\unit{N}$. The tuning parameters of the path planner are $w_X=0.1$, $w_I=1$, $w_q=0.5$, $w_u=1$, $w_z=50$ and $N=8$ and time of prediction $T$ is 0.1~$\unit{s}$. The constraints on altitude, attitude and velocity of the UAV are presented in~\eqref{eq:bounds}. In all the following cases the agent's initial condition is $X_0 = [1,\,0,\,-0.8,\,0,\,0,\,0]$ and it is assumed that the downward camera has a VFOV and a HFOV of 1.2~$\unit{rad}$.
\begin{subequations} \label{eq:bounds}
\begin{align}
0 \le &z_k \le \unit[1]{m} \\
-\nicefrac{\pi}{10} \le &\phi_k \le \unit[\nicefrac{\pi}{10}]{rad} \\
-\nicefrac{\pi}{10} \le &\theta_k \le \unit[\nicefrac{\pi}{10}]{rad} \\
-\pi \le &\psi_k \le \unit[\pi]{rad} \\
-2 \le &\dot{x}_k \le \unitfrac[2]{m}{s} \\
-2 \le &\dot{y}_k \le \unitfrac[2]{m}{s} \\
-2 \le &\dot{z}_k \le \unitfrac[2]{m}{s}
\end{align}
\end{subequations}

In order to reduce the computation time, the target area is filled with random points and the points inside of each cell $c_k$ are enumerated, providing an estimate of the actual area coverage. The average computation time at each step for all scenarios is $\unit[0.5]{s}$ and in the vast majority of tests the runtime is below $\unit[1]{s}$. Additionally, it is assumed that a low-level attitude controller is able to track the desired roll, pitch, yaw and thrust. Additive noise with standard deviation of 3$\unit{cm}$ for the positioning system and of $\pm1$~$\unit{N}$ for the thrust are considered, to simulate the effect of wind gust or other possible external disturbances. In what follows, different scenarios are presented in order to evaluate the performance of the proposed method. All simulations have been performed on a computer with an Intel Core i7-6600U CPU, 2.6GHz and 8GB RAM. For visualization purposes and without loss of generality the orientation of the camera in the figures was down sampled.

\begin{figure}[htbp!]
\centering
	\includegraphics[width=\linewidth]{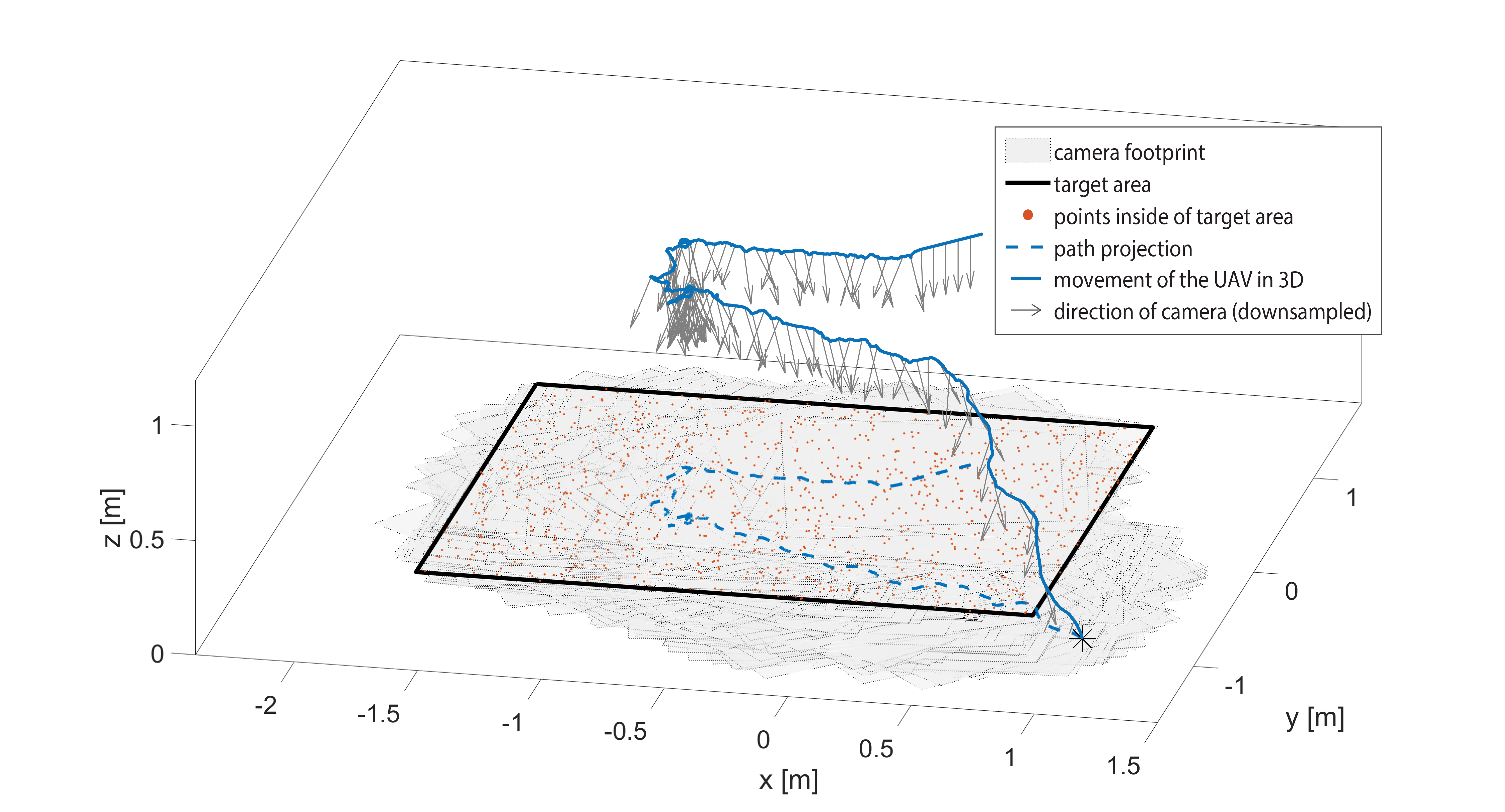}
        \caption{(Case 1) Motion of the UAV with the proposed particle harvesting methodology for visual area coverage. The target area is circumscribed by a thick black line. The camera footprints are shown as light gray rectangles. The projection of the path on the ground is shown with a dashed blue line.}
        \label{fig:3d2drec}
 \end{figure}

In the first scenario, the target area is considered to be a rectangle with a size of $2.5 \times 2$\,$\unit{m}$. The generated path is depicted in Fig.~\ref{fig:3d2drec}. Moreover the camera footprints of the motion in relation to the target area are shown. It can be seen that despite the fact the UAV is inside the target area, the camera captures a larger area based on the UAV's attitude. This shows the importance of considering the time-varying camera's footprint in path planner algorithms. 



\begin{figure}[htbp!]
\centering
        \includegraphics[width=\linewidth]{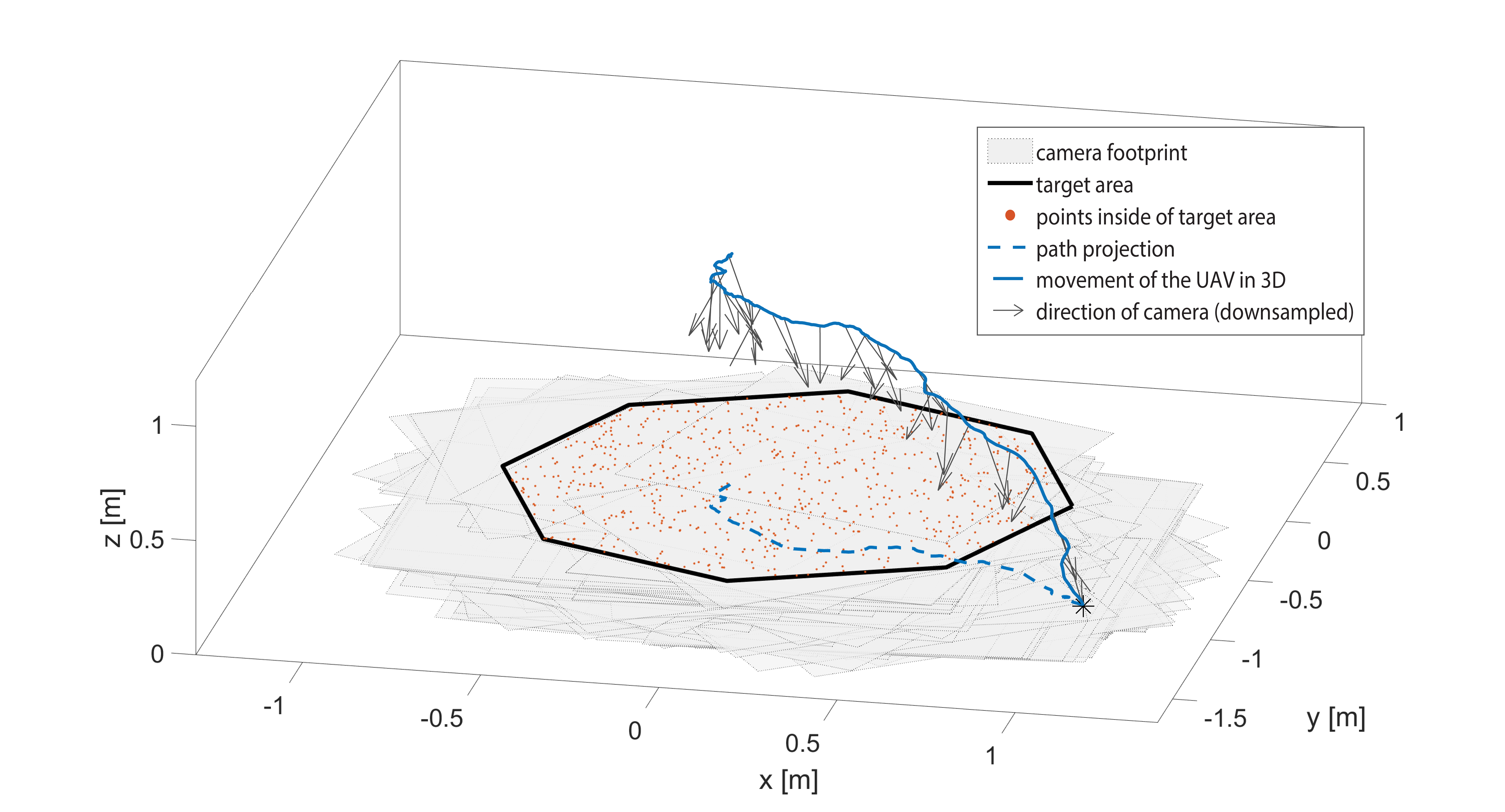}
        \caption{(Case 2) Motion of the UAV with the proposed particle harvesting methodology for visual area coverage.The target area and camera footprints are shown.}
        \label{fig:3d2dOct}
\end{figure}

In the second case, an octagonal polygon with an area of $1.8$~$\unit{m^2}$ is considered. The result of the path planner's trajectory and footprints of the camera are depicted in Fig.~\ref{fig:3d2dOct}. The area is covered completely and there is sufficient overlap between the camera footprints, which is critical for further image post processing~\cite{brown2007automatic}.

\begin{figure}[htbp!]
\centering
        \includegraphics[width=\linewidth]{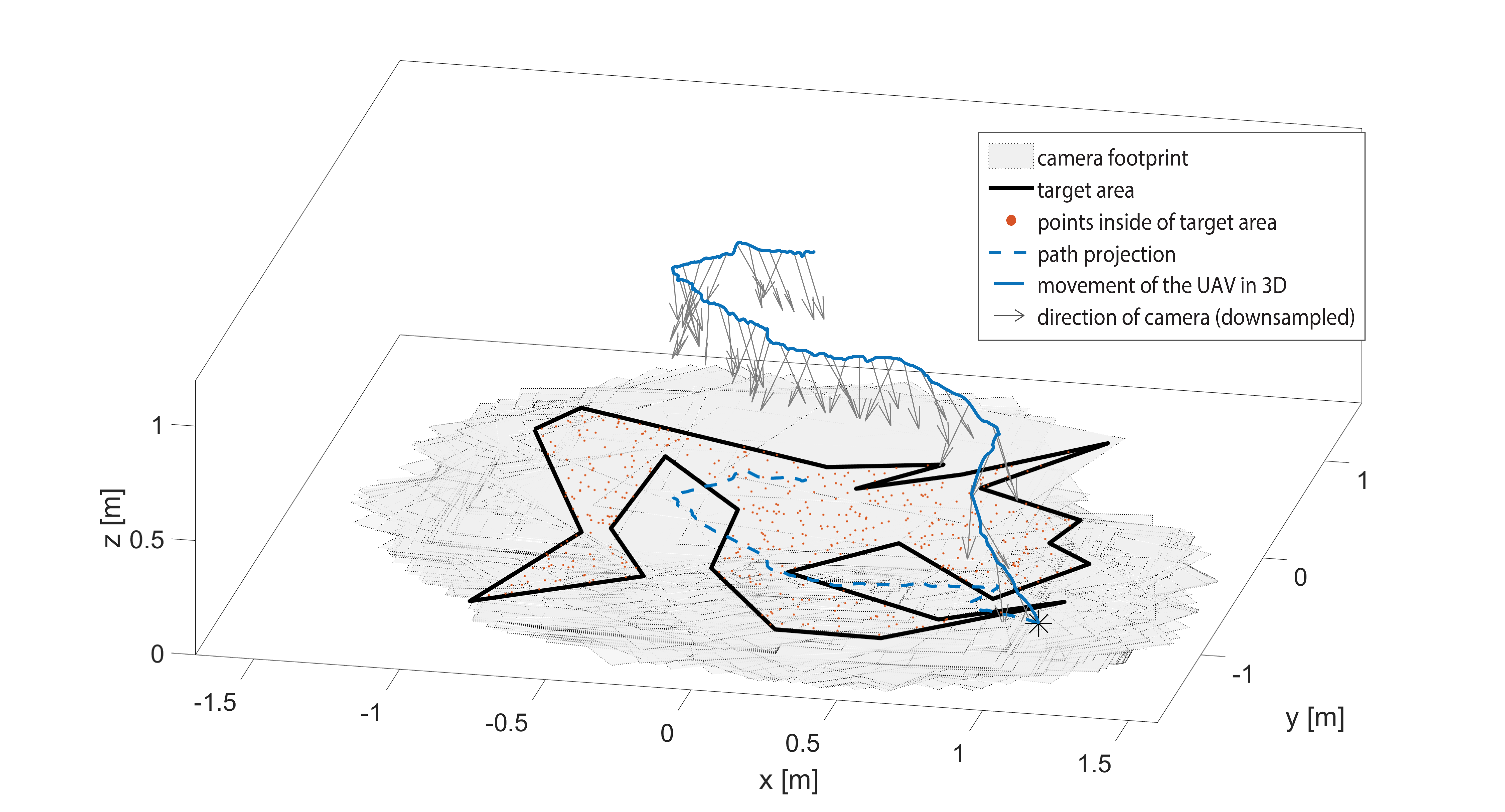}
        \caption{(Case 3) Motion of the UAV with the proposed particle harvesting methodology for visual area coverage.The target area and camera footprints are shown.}
        \label{fig:3d2dnonconv}
\end{figure}

Furthermore, Fig.~\ref{fig:3d2dnonconv} presents the obtained results in the case of a complex nonconvex polygon. In this case the area is fully covered by the path planner. Additionally, it should be highlighted that, in all scenarios, the UAV does not sweep the area to achieve full coverage; by considering the camera footprints in the path planner, the path becomes reasonably smooth and shorter as there is no need to sweep the area for full coverage. As an example, it can be seen in Fig.~\ref{fig:3d2dOct} that the UAV covered all of the area, while simply moving on a rather simple path towards the center of the domain of the target area.

\begin{figure}[htbp!]
\centering
                \includegraphics[width=\linewidth]{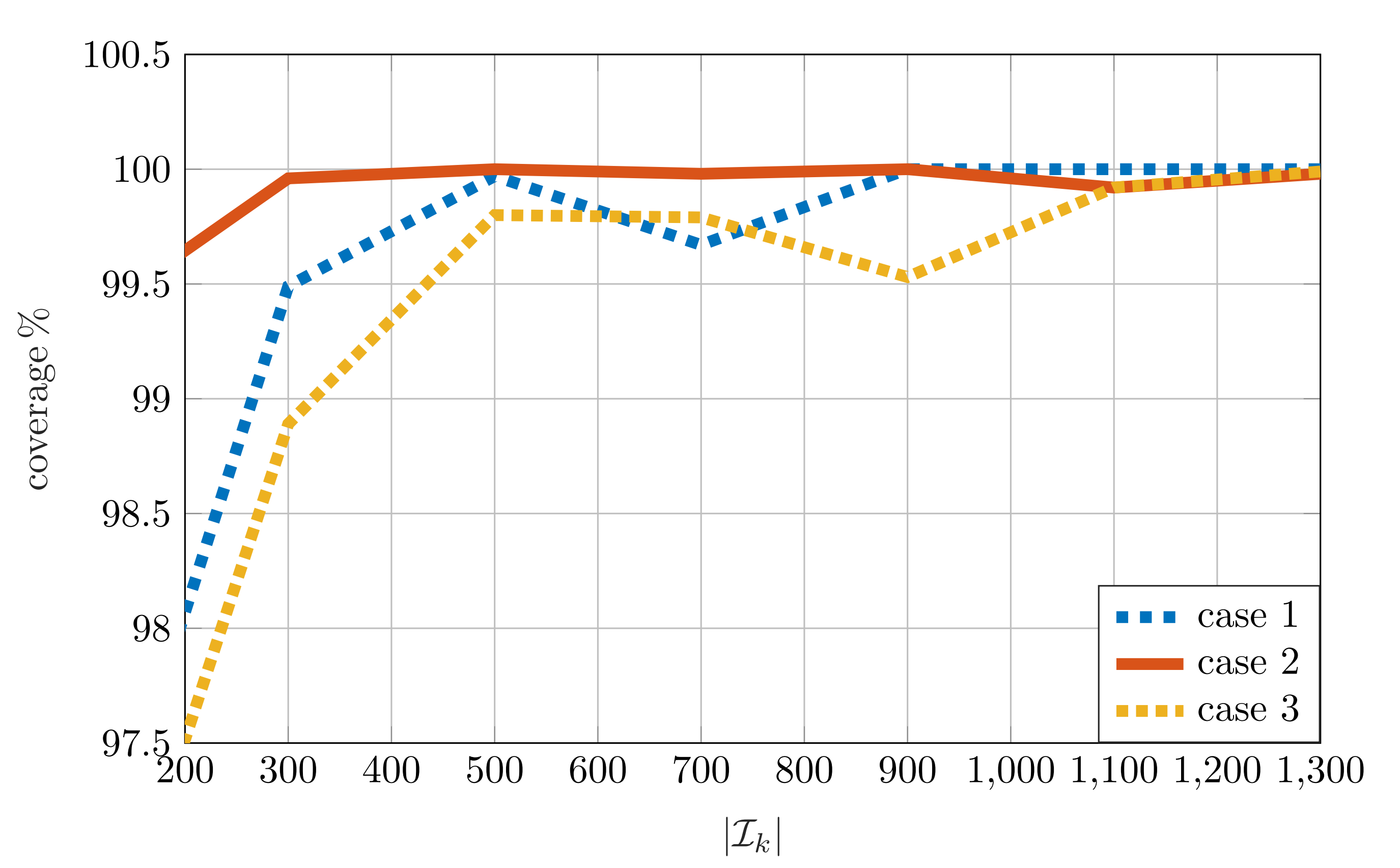}
        \caption{Relation between the coverage percentage and the number of random points (particles). The coverage percentage is the ratio between the covered area over the overall area. Note that even with a low density of particles, we achieve coverage levels above $\unit[97.5]{\%}$}
        \label{fig:comparison}
\end{figure}

\begin{figure}[htbp!]
\centering
    \includegraphics[width=\linewidth]{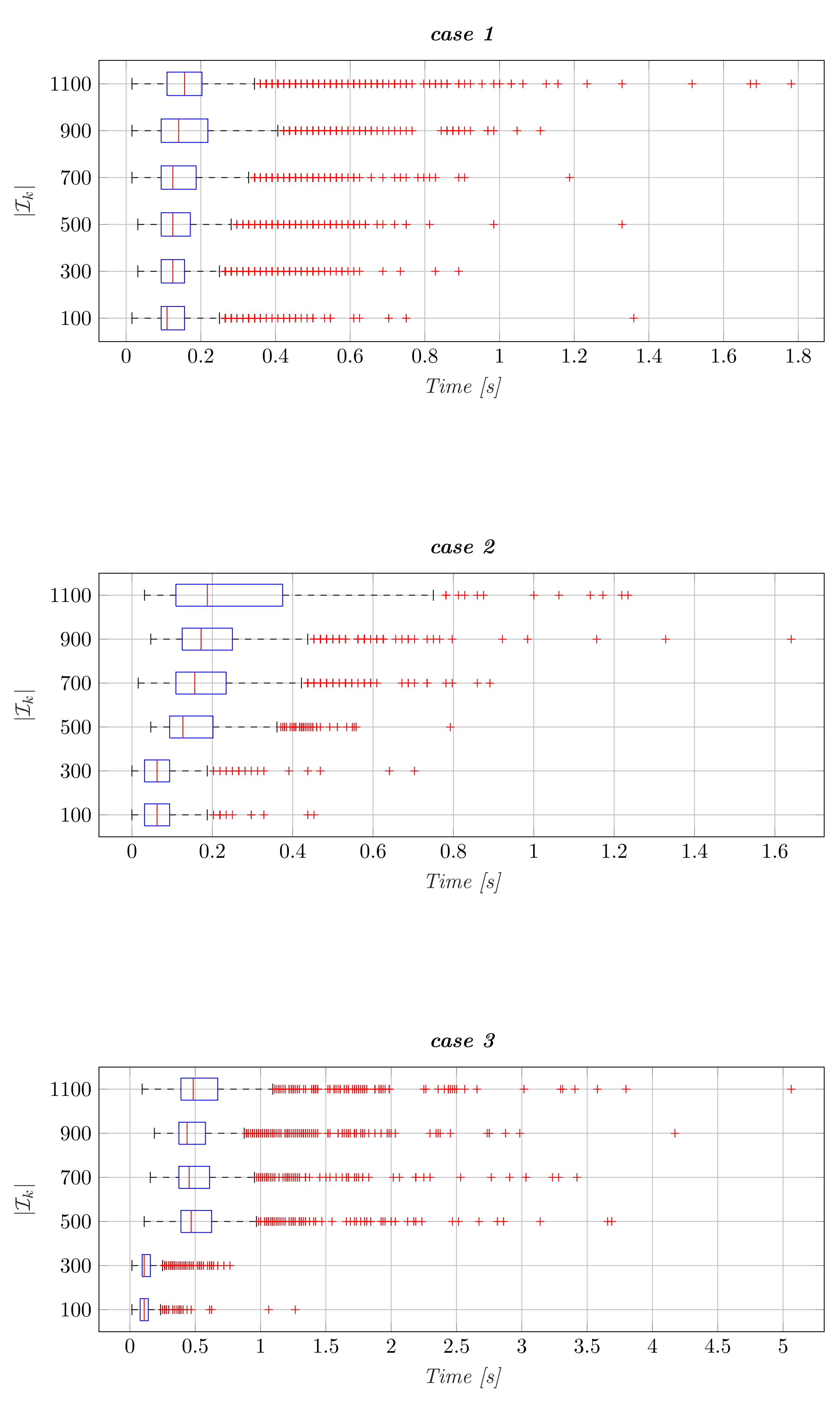}
        \caption{Computation time of each case in relation to the number of random points.}
        \label{fig:comparisontimevspoints}
\end{figure}

Moreover, Table~\ref{table:pathplanner} summarizes the path length and compares the proposed path planner, the path generates by the grid based method and the method in~\cite{MANSOURI20181}. The flight time depends on the technical characteristics of the UAV, whereas the total length of the path is more appropriate to be used for comparisons. In the benchmark methods, the UAV maneuvers at a fixed altitude of 1~$\unit{m}$. In case of grid based method the UAV sweeps the grids with a minimum number of turns to cover the target area and the method in~\cite{MANSOURI20181} segments the area with fixed size camera footprint and provides shortest path that pass through all camera footprints. It is shown that the proposed path planner significantly reduces the length of the path in all cases, while the increase in prediction horizon $N$ results in shorter paths. This shows the impact of the time-varying footprint in path planners. 
{\renewcommand{\arraystretch}{1.2}
\begin{table}[htbp!]
\centering
\caption{The comparison of the path length between the proposed path planner and the path from grid based method and method~\cite{MANSOURI20181}.}
\label{table:pathplanner}
\begin{tabular}{cccc}
\cline{2-4}                                                         & Case 1 & Case 2 & Case 3 \\ \hline
\multicolumn{1}{c} {\begin{tabular}[c]{@{}c@{}}path from grid-based method \end{tabular}} & \unit[7.1]{m}    & \unit[4.2]{m}    & \unit[8.1]{m}    \\ \hline
\multicolumn{1}{c}{\begin{tabular}[c]{@{}c@{}}path from method~\cite{MANSOURI20181} \end{tabular}}   & \unit[7.1]{m}   & \unit[3.6]{m}  & \unit[6.9]{m}   \\ \hline
\multicolumn{1}{c}{proposed path planner with $N=8$}                         & \textbf{\unit[5.6]{m}}    & \textbf{\unit[2.3]{m}}   & \textbf{\unit[3.5]{m}}  \\ \hline
\multicolumn{1}{c}{proposed path planner with $N=15$}                         & \textbf{\unit[4.8]{m}}    & \textbf{\unit[1.7]{m}}   & \textbf{\unit[2.8]{m}}  \\ \hline
\end{tabular}
\end{table}
}\\
Furthermore, Fig.~\ref{fig:comparison} depicts the performance of the proposed path planner in relation to the number of random points inside of the target area. It is shown that the proposed path planner performance increases by the number of particles, however, more points can affect the computation time. Figure~\ref{fig:comparisontimevspoints} depicts that the average and maximum value of computation time change in relation to the number of random points. Thus, more investigation is needed in order to determine the optimal number of random particles.

\section{Conclusions} \label{conclusion}
In this article we presented the particle harvesting methodology: a novel approach to the problem of visual area coverage and path planning with attitude-dependent footprints. The path planner takes into account the kinematics of the UAV, the image quality and the camera footprint to generate attitude references for the low-level controller. This way, the segmentation of the area and the path planner are integrated and the UAV does not require to sweep the area for complete coverage. This leads to shorter paths compared to other methods as shown in Table~\ref{table:pathplanner}, which is an important advantage due to the limited flight time of the UAVs. The navigation problem using the proposed methodology is solved at every time instant in a receding horizon fashion. 

Furthermore, the path planner is adequately fast which establishes an overall framework for the path planning of the UAV with a downward camera in order to solve the coverage problem. The presented method has been tested in different convex and non-convex polygons and in all examined scenarios, a highly satisfactory coverage has been obtained. 
The coverage is found to be consistently close to $\unit[100]{\%}$ in all experiments. In order to achieve $\unit[100]{\%}$ coverage, the original area can be slightly enlarged. The provided path for the UAV leads to frames with substantial overlapping, which is necessary for the algorithm to provide a visual overview of the scene. 

Future work will focus on the investigation of the effect of multiple UAVs, time-varying areas, collision avoidance, as well as the extensive experimentation with the overall suggested scheme, including the optimal number of particles generated inside the area of interest.

\section*{References}
\bibliography{mybib}
\end{document}